\documentclass[journal]{IEEEtran}
\ifCLASSINFOpdf
\else
\fi
\usepackage{lineno,hyperref}
\modulolinenumbers[5]
\usepackage{graphicx}
\usepackage{epstopdf}
\usepackage{subfigure}
\usepackage{amssymb,amsmath}
\usepackage{multirow}
\usepackage{multicol}
\usepackage{array}
\usepackage{amsmath}

\usepackage{times}
\usepackage{epsfig}
\usepackage{graphicx}
\usepackage{amsmath}
\usepackage{amssymb}
\usepackage{subfigure}
\usepackage{color}
\usepackage[ruled]{algorithm2e}
\usepackage{ragged2e}
\usepackage{array}

\begin{document}
\title{A Semantics-Guided Class Imbalance Learning Model for Zero-Shot Classification}

\author{Zhong~Ji,~\IEEEmembership{Member,~IEEE,}
        Xuejie~Yu,
        Yunlong~Yu*,
        Yanwei~Pang,~\IEEEmembership{Senior Member,~IEEE,}
        and~Zhongfei~Zhang% <-this % stops a space

\thanks{This work was supported by the National Natural Science Foundation of
China under Grants 61771329.}%
\thanks{Z.~Ji, X.~Yu, Y.~Yu* (corresponding author) and Y.~Pang are with the School of Electrical and Information Engineering, Tianjin University, Tianjin 300072, China (e-mail: \{jizhong, xuejie\_yu,~yuyunlong,~pyw\}@tju.edu.cn).}% <-this % stops a space
\thanks{Z. Zhang is with the Computer Science Department, Watson School, the State University of New York Binghamton University, Binghamton, NY 13902, USA (e-mail: zhongfei@cs.binghamton.edu).}
}

\maketitle

\begin{abstract}
Zero-Shot Classification (ZSC) equips the learned model with the ability to recognize the visual instances from the novel classes via constructing the interactions between the visual and the semantic modalities. In contrast to the traditional image classification, ZSC is easily suffered from the class-imbalance issue since it is more concerned with the class-level knowledge transfer capability. In the real world, the class samples follow a long-tailed distribution, and the discriminative information in the sample-scarce seen classes is hard to be transferred to the related unseen classes in the traditional batch-based training manner, which degrades the overall generalization ability a lot. Towards alleviating the class imbalance issue in ZSC, we propose a sample-balanced training process to encourage all training classes to contribute equally to the learned model. Specifically, we randomly select the same number of images from each class across all training classes to form a training batch to ensure that the sample-scarce classes contribute equally as those classes with sufficient samples during each iteration. Considering that the instances from the same class differ in class representativeness, we further develop an efficient semantics-guided feature fusion model to obtain discriminative class visual prototype for the following visual-semantic interaction process via distributing different weights to the selected samples based on their class representativeness. Extensive experiments on three imbalanced ZSC benchmark datasets for both the Traditional ZSC (TZSC) and the Generalized ZSC (GZSC) tasks demonstrate our approach achieves promising results especially for the unseen categories those are closely related to the sample-scarce seen categories.

\end{abstract}


% \fi
\IEEEpeerreviewmaketitle

\section{Introduction}

\IEEEPARstart{O}{bject} classification has made remarkable progress with the emergence of deep learning \cite{szegedy2015going,yu2018transductive,simonyan2014very} and large-scale datasets \cite{krizhevsky2012imagenet,he2016deep,wu2018deep}. However, the traditional supervised models are data-hungry that require a large amount of well-labeled training data to feed them up and are unable to generalize to new categories.
Inspired by humans' ability to recognize the objects from new categories at the first sight with only some semantic descriptions, Zero-Shot Classification (ZSC) \cite{IEEEpami:Lampert,guo2017zero,IEEEcvpr15:Akata,ICLR14:Norouzi,IEEEnips15:Romera-Paredes,bulent2019gradient} gains huge popularity recently. ZSC aims at handling the unseen categories absent from the training phase based on some auxiliary semantic information, e.g., user-defined attributes \cite{IEEEpami:Lampert,IEEEcvpr09:Farhadi}, word vectors of the class names \cite{IEEEcvpr15:Akata,morgado2017semantically,IEEEnips13:Mikolov} and the text descriptions \cite{lei2015predicting,qiao2016less,elhoseiny2017link,zhu2018generative}. It is typically achieved by resorting to constructing the interactions between the visual and the semantic modalities in a class semantic embedding space.

In real world classification scenarios, the numbers of images per class are always subject to a long-tailed distribution \cite{Technical11:Wah}. For the common categories like ``cat'' and ``dog'', it is relatively easy to obtain sufficient samples. However, collecting enough samples for certain categories such as ``beaver'' and ``mole'' is hard due to the species scarcity or technical problems. Therefore, it is quite difficult to build a large-scale dataset with a balanced sample size  for each class. How to enable the learned model to fit the classes with a small number of samples is worth to be addressed \cite{zhang2016transfer,lim2016evolutionary}. Unfortunately, the class imbalance issue has been far neglected in ZSC. 
\begin{figure}[t]
	\centering
	\includegraphics[scale=0.6]{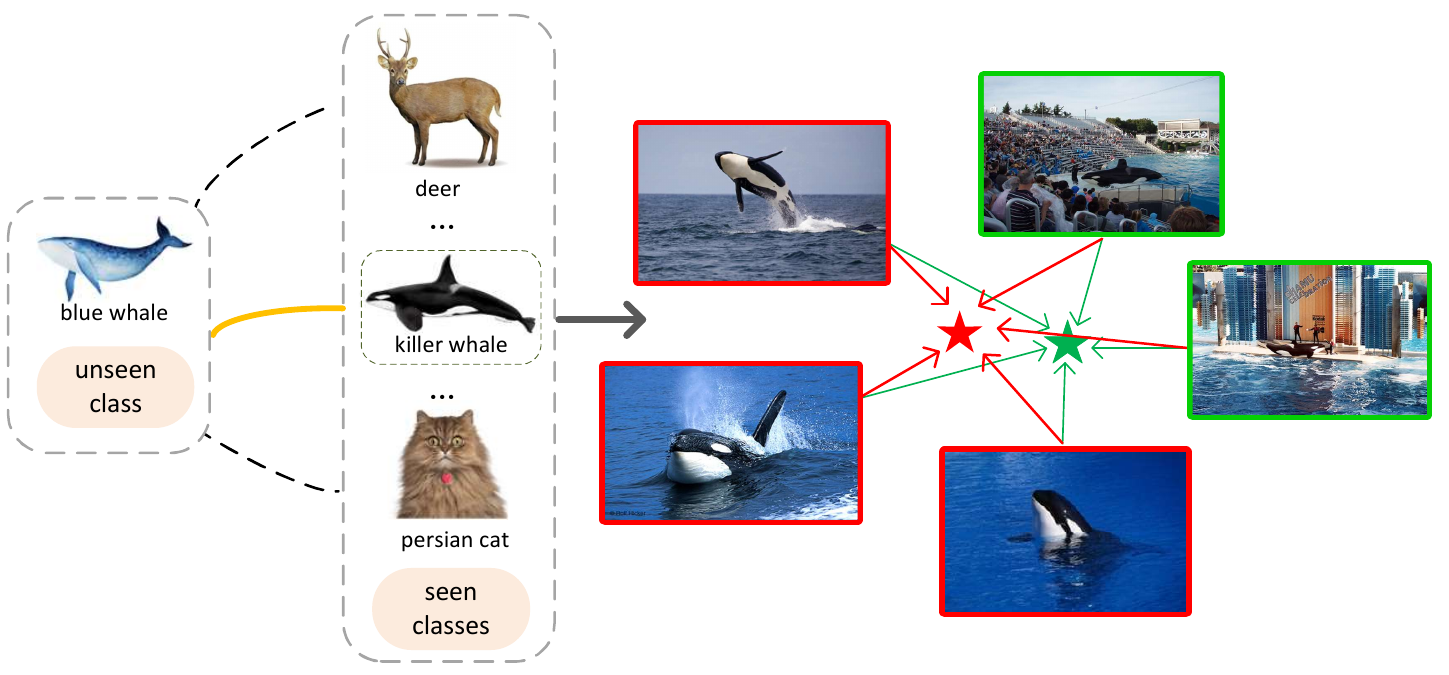}
	\caption{An illustration of the basic motivation of the proposed method. In the class-level, the transfer information for the unseen class ``blue whale'' mainly comes from its affinity seen class ``killer whale'' (denoted in yellow solid line), while the unrelated classes such as ``deer'' and ``persian cat'' contribute little to the knowledge transfer (denoted in black dotted line). In the instance-level, the intra-class instances differ in their representativeness. Two strategies are provided to obtain the class visual prototype based on with (red line) or without (green line) considering the representativeness differences of intra-class instances. Best view in color.
	}
\label{fig1}
\end{figure}

Compared with the traditional object recognition models that have no new categories in the test stage, ZSC is more concerned with the knowledge transfer capability on class-level. Since the training and test categories are non-overlap but related in ZSC setting, the seen categories closely related to a certain target unseen category require to be well trained to facilitate the knowledge transfer. Most of the recent ZSC approaches are based on deep learning technique and trained with the stochastic gradient descent (SGD) in a mini-batch way. Obviously, if the seen categories have a small number of samples, this batch-based training fashion is hard to ensure the generalized ability to the affinity unseen categories. On one hand, during each batch iteration, the samples from the sample-scarce categories have lower probability to be selected, resulting in the key discriminant information unable to be transferred to the closely related unseen categories. On the other hand, the learned model will have a strong bias on the categories with sufficient data, leading to the overfitting problem and largely degrading the average generalization ability for all unseen categories.
As illustrated in Fig.~\ref{fig1}, the discriminant information to recognize the unseen category ``blue whale" is mainly transferred from its affinity seen category ``killer whale", while the unrelated seen categories such as ``deer" and ``persian cat" hardly provide transferable information. If the seen category ``killer whale" has a small number of samples, the model would generalize poorly on the unseen category ``blue whale".

\begin{figure*}[t]
	\centering
	\includegraphics[scale=1.14]{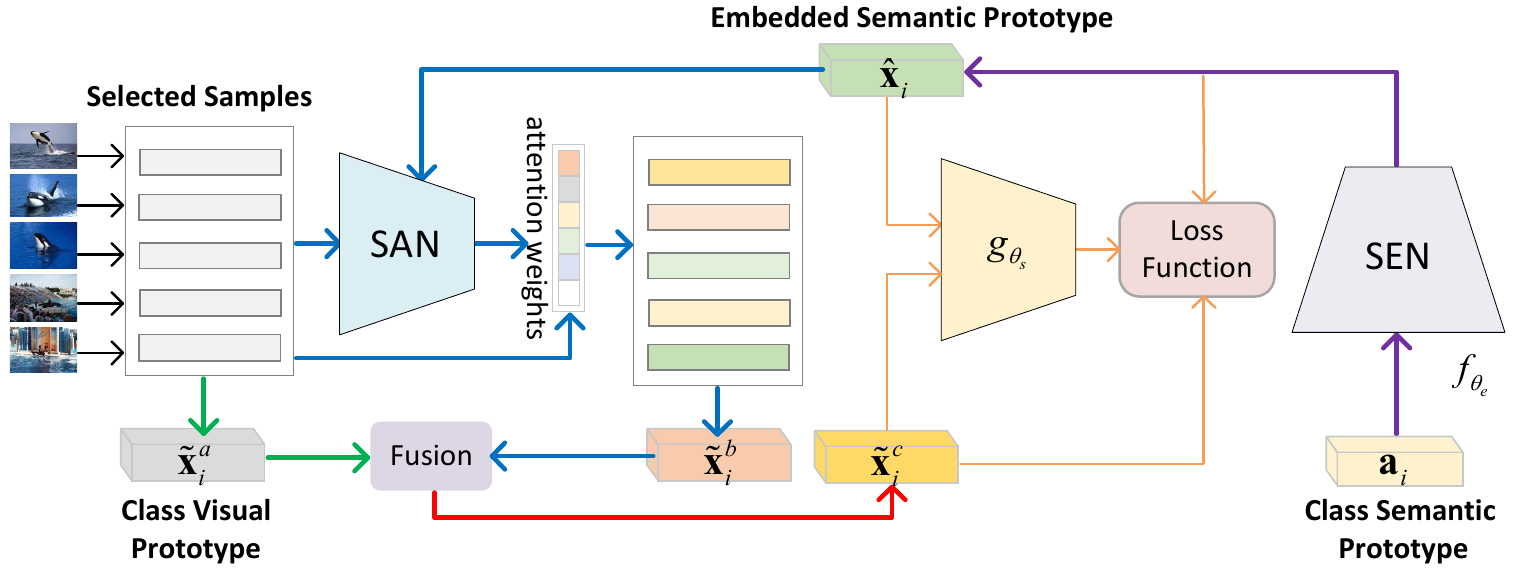}
	\caption{The Semantics-Guided Class Imbalance Learning Model (SCILM). For the $i$-th class, SCILM first embeds the class semantic feature into the visual feature space to align visual-semantic modalities and obtain the embedded semantic prototype ${{\bf{\hat x}}_i}$ via SEN (denoted in purple line). Then, SCILM extracts three different class visual prototypes ${{\bf{\tilde x}}_i}$, denoted as ${{\bf{\tilde x}}_i^a}$, ${{\bf{\tilde x}}_i^b}$, and ${{\bf{\tilde x}}_i^c}$, respectively. The first one ${{\bf{\tilde x}}_i^a}$ is obtained by directly averaging the visual features of the selected instances (denoted in green line). The second one ${{\bf{\tilde x}}_i^b}$ is guided by weighting the representativeness of each selected samples via SAN (denoted in blue line), thus encouraging the more discriminative samples and suppressing the noisy samples. The third one ${{\bf{\tilde x}}_i^c}$ is obtained by an effective fusion measure to make full use of the complementary information in ${{\bf{\tilde x}}_i^a}$ and ${{\bf{\tilde x}}_i^b}$. The final loss function is derived from ${{\bf{\hat x}}_i}$ and ${{\bf{\tilde x}}_i}$, which is denoted in yellow line and explained in detail in Section \uppercase\expandafter{\romannumeral3}.}
\label{fig2}
\end{figure*}

In order to mitigate the limitation of the traditional batch-based training methods on the class imbalance issue, we propose a novel training strategy to balance the training model by randomly selecting the same number of samples from each of training classes to form a class-balanced batch during each iteration to ensure that the sample-scarce categories have the same decision effect as the sample-abundant categories. Based on this training strategy, we further propose an efficient model to synthesize the class visual prototype via distributing different weights for the selected instances based on their class representativeness, which is achieved with a Semantic Embedding Network (SEN) and a Semantic Attention Network (SAN), respectively. Specifically, the SEN module embeds the class semantic prototypes into the visual space so as to encode the semantic relationships between the visual instances and the class semantics. The SAN module assigns representativeness weight to each selected sample according to its category contribution represented by the semantic similarity between the visual instance and the corresponding class semantic prototype. The learned model is guided to suppress the noisy background (e.g., the images with green bounding box in Fig. \ref{fig1}) and focus on the semantically more related images (e.g., the images with red bounding box in Fig. \ref{fig1}), yielding a more discriminative and robust visual prototype. The whole approach is referred as Semantics-Guided Class Imbalance Learning Model (SCILM) and its flowchart is shown in Fig.~\ref{fig2}.

%The AFEM module is explained in detail in Section \uppercase\expandafter{\romannumeral3} and
%Then, an Average Feature Extraction Model (AFEM) module is exploited to extract the class visual prototypes, which is the core of our proposed method. AFEM is consisted of two main parts, i.e., the Semantic Embedding Network (SEN) and the Semantic Attention Network (SAN). The SEN component embeds the

To summarize, the main contributions of this work are as follows:
\begin{enumerate}
	\item We provide a novel class-balanced learning model to alleviate the class imbalance issue in ZSC. To the best of our knowledge, we are the first to cope with ZSC from the perspective of class imbalance problem. Specifically, instead of randomly selecting a batch-size of images from all training data, we randomly select the same number of images from each class across all training classes to form a training batch during each iteration. In this way, the sample-scarce classes are encouraged to have the same decision effect and contribute equally with the multi-sample classes.
	\item In order to learn a more general semantics-visual interaction model, we propose to embed the class semantics into the visual space under the supervision of the class visual prototypes, i.e., the fusion of the selected visual instances from each class. To obtain a more discriminative visual prototype, we further develop a semantics-guided feature fusion model to distinguish the representativeness of the visual instances based on their relevances to the class semantic prototypes. The overall model is forced to attend on semantically more related samples and suppress the interference of background noise and outlier images.

	\item Extensive experiments on three ZSC benchmark datasets that easily suffer from the class imbalance issue, i.e., AwA1 \cite{IEEEpami:Lampert}, AwA2 \cite{IEEEcvpr17:Xian}, and aPY \cite{IEEEcvpr09:Farhadi}, demonstrate the effectiveness of our method. For the Traditional ZSC (TZSC) \cite{IEEEcvpr17:Xian} task, our proposed approach obtains comparable performances compared with the state-of-the-art methods and increases the baseline method DEM \cite{IEEEcvpr17:Zhang} by 4.1\%, 3.0\%, and 3.4\%, respectively. For the Generalized ZSC (GZSC) task \cite{IEEEcvpr17:Xian}, the \textbf{H} metric is improved by 1.5\%, 1.8\%, and 5.3\% against the state-of-the-art methods, respectively.
\end{enumerate}

The remaining sections of the paper are organized as follows. Section \uppercase\expandafter{\romannumeral2} reviews the related work. Section \uppercase\expandafter{\romannumeral3} introduces our proposed SCILM method in detail. Section \uppercase\expandafter{\romannumeral4} presents the experiments and analyses, followed by the conclusion in Section \uppercase\expandafter{\romannumeral5}.

\section{Related Work}
ZSC tasks are typically addressed by modeling the interactions between visual and semantic modalities in a linear or deep learning based way. In this section, we roughly divide the related work into two parts, i.e., the linear based methods and the deep learning based methods. Besides, we have a brief review about the Generalized Zero-Shot Classification (GZSC).
\subsection{Linear Based Methods}
Early ZSC studies follow the paradigm to formulate the model with a linear matrix constrained by different loss functions. Among them, DAP \cite{IEEEpami:Lampert} learns an attribute classifier by maximizing the attribute posteriors. To further employ the intra-class semantic relationships, ALE \cite{akata2015label} employs a bilinear embedding model with a hinge ranking loss. ESZSL \cite{IEEEnips15:Romera-Paredes}, SJE \cite{IEEEcvpr15:Akata} and LATEM \cite{IEEEcvpr16:Xian} explore a joint embedding space to maximize the cross-modal compatibility scores. ConSE \cite{ICLR14:Norouzi} utilizes the classifier scores as weights to form the semantic space spanned by the convex combination of semantic features to predict the class posteriors.  SynC \cite{IEEEcvpr16:Changpinyo} synthesizes classifiers for unseen categories based on phantom prototypes via aligning the semantic space and model space. SAE \cite{IEEEcvpr17:Kodirov} uses a linear and symmetric auto-encoder approach to cope with the domain shift issue in ZSC. Following the encoder-decoder paradigm, LSE \cite{Yu2018zero} encodes different modalities in a feature aware latent space to maximize the recoverability and predictability of the feature space and latent space, respectively. SSE \cite{iccv15:Zhang} represents each class as a probabilistic mixture of seen classes and forces the images from the same class to have similar mixture patterns. CDL \cite{jiang2018learning} proposes a coupled dictionary learning model to improve the incomplete semantic space with the discriminative property lying in the visual space. JEDM \cite{yu2018transductive} also develops a dictionary approach but solves ZSC in a self-training way and uses transductive approach to deal with the domain shift issue. BZSL \cite{shen2019scalable} models the visual instances and class embeddings with two binary functions in an intermediate Hamming space.

\subsection{Deep Learning Based Methods}
Early methods applying deep metrics to ZSC mainly focus on learning a visual-semantic embedding by resorting to the neural networks and are trained with stochastic gradient descent (SGD) in a mini-batch way. Among them, DeViSE \cite{IEEEnips13:Frome} and CMT \cite{IEEEnips13:Socher} follow the visual-semantic embedding direction, that is, leverage a deep neural network to embed the visual features into the semantic space. In contrast, some approaches \cite{IEEEiccv15:Kodirov,ECML15:Shigeto,IEEEcvpr17:Zhang,zhao2017zero,wang2016relational} formulate an inverse mapping direction from the semantic space to the visual space, i.e., semantic-visual embedding. It has been experimentally proved that the semantic-visual embedding is able to generate more compact and separative visual feature distribution with the one-to-many correspondence manner, thereby mitigating the hubness issue \cite{ECML15:Shigeto}. More related to our work is the DEM \cite{IEEEcvpr17:Zhang} model, which employs the visual space as embedding space and is optimized in an end-to-end manner. We adopt the similar ridge regression loss function and the neural network based deep embedding model to project the semantic features into the visual feature space. Note that our SCILM model focuses on the class imbalance issue without introducing additional data, which is ignored in DEM \cite{IEEEcvpr17:Zhang} and other deep or linear models.

To further explore the data distribution information in the embedding space, more recently, many deep learning based work aims to generate pseudo visual features taking as input the class semantic prototypes. From this perspective, the semantic-visual embedding models are reasonably regarded as a semantically supervised feature generation process. The other branch \cite{IEEEcvpr18:Zhu,felix2018multi,chen2018zero,zhang2018sch, bulent2019gradient} of deep feature generation methods is to generate samples with Generative Adversarial Network (GAN) \cite{goodfellow2014generative}, which approximates the distributions of unseen classes from an instance. Among them, GAZSL \cite{IEEEcvpr18:Zhu} takes as input the noisy semantic text descriptions and generates unseen visual features. Cycle-CLSWGAN \cite{felix2018multi} forces the generated visual feature to reconstruct the corresponding semantic feature in a multi-modal cycle-consistent manner. SP-AEN \cite{chen2018zero} tackles the semantic loss problem in ZSC by introducing two independent semantic embeddings with adversarial learning.

However, both the linear and deep learning based models are not specially designed to well generalize for the sample-scarce categories. The learned matrices in the linear models are inclined to the class with sufficient samples. For the deep learning based methods trained in a mini-batch way, the overall generalization ability to novel classes is more vulnerable for the training classes with small samples. On the contrary, our proposed method encodes discriminative class-level visual prototypes guided by the semantic information to balance the model for the classes with a small number of images. Besides, compared with the GAN-based deep methods, SCILM has a much simpler framework without complex adversarial training, ensuring the stability and fast convergence.

\subsection{Generalized Zero-Shot Classification (GZSC)}
% SynC []   3ME [] DEM []
In the Traditional Zero-Shot Classification (TZSC) setting, the model is trained with the seen visual features and the shared attributes annotated on class-level, and tested solely on the unseen classes, which is limited on the strong assumption that the test images only come from the novel classes. The Generalized Zero-Shot Classification (GZSC) \cite{IEEEcvpr17:Xian,chao2016empirical} promotes ZSC tasks to a more realistic scenario, that is, the test set is opened to all classes (the seen and unseen classes). It measures the model's overall robustness by the harmonic mean of the prediction accuracy on seen and unseen classes, and obtains optimal performance when both seen and unseen images are correctly classified as possible.

\section{Methodology}

\subsection{Notations}
We assume $\{ ({{\bf{x}}_i},{{\bf{y}}_i},{{\bf{a}}_i})|{{\bf{x}}_i} \in {{\bf{X}}_S},{{\bf{y}}_i} \in {{\bf{Y}}_S},{{\bf{a}}_i} \in {{\bf{A}}_S}\}$ to denote the training (seen) set data. ${{\bf{X}}_S} = \{ {{\bf{x}}_1},...,{{\bf{x}}_i},...,{{\bf{x}}_N}\}  \in {\mathbb{R}^{N \times p}}$ represents the image visual feature set, where $N$ and $p$ denote the number and the dimensionality of visual features, respectively. ${{\bf{Y}}_S} = \{ {{\bf{y}}_1},...,{{\bf{y}}_i},...,{{\bf{y}}_k}\}  \in {\mathbb{R}^{N \times k}}$ denotes the one-hot label set of $k$ training (seen) classes. ${{\bf{A}}_S} = \{ {{\bf{a}}_1},...,{{\bf{a}}_i},...,{{\bf{a}}_k}\}  \in {\mathbb{R}^{k \times q}}$ refers to the training (seen) class semantic feature set, where $q$ is the dimensionality of semantic features. Let $\{ ({{\bf{x}}_u},{{\bf{y}}_u},{{\bf{a}}_u})|{{\bf{x}}_u} \in {{\bf{X}}_U},{{\bf{y}}_u} \in {{\bf{Y}}_U},{{\bf{a}}_u} \in {{\bf{A}}_U}\}$ represent the test (unseen) data set from $t$ test (unseen) classes. Note that ${{\bf{Y}}_S} \cap {{\bf{Y}}_U} = \varnothing $ in the ZSC setting. Given a test sample ${{\bf{x}}_u} \in {{\bf{X}}_U}$, the task of Traditional ZSC (TZSC) is to classify it into the unseen label space, while the Generalized ZSC (GZSC) classifies it into the joint label space spanned by both seen and unseen labels.

\subsection{The Semantics-Guided Class Imbalance Learning Model (SCILM)}
The key to our proposed method is to obtain the class visual prototype guided by the class semantic feature via a Semantics-Guided Class Imbalance Learning Model (SCILM). As illustrated in Fig. \ref{fig2}, SCILM module is mainly composed of two parts: the Semantic Embedding Network (SEN) and the Semantic Attention Network (SAN), which will be introduced detailedly in the following.
\paragraph{\textbf{Semantic Embedding Network (SEN)}} In order to establish the connection between the class semantic modality and the visual modality, we need to align them in a common space. Here, SEN module projects the class semantic feature into the visual feature space considering the benefit of alleviating the hubness problem \cite{ECML15:Shigeto}. Specifically, as seen in Fig. \ref{fig2} (denoted in purple line), we map the semantic feature into the visual feature space with a three-layer embedding neural network to get the class semantic prototype, which is formulated as follows:
\begin{equation}
{{\bf{\hat x}}_i} = {f_{{\theta _{e}}}}({{\bf{a}}_i}) = \delta ({{\bf{W}}_2} \cdot \delta ({{\bf{W}}_1} \cdot {{\bf{a}}_i} + {{\bf{b}}_1}) + {{\bf{b}}_2}),
\end{equation}
where $\mathbf{a}_i$ is the class-level semantic feature input for $i$-th class, $\mathbf{\hat x}_i$ is the class semantic embedding in the visual space, $\delta$ denotes the ReLU \cite{nair2010rectified} activation function and ${f_{{\theta _{e}}}}( \cdot )$ represents the mapping function of SEN. ${\theta _{e}}$ represents the trained parameter set, which includes $\{\mathbf{W}_1, \mathbf{W}_2, \mathbf{b}_1, \mathbf{b}_2\}$.

\paragraph{$\textbf{Semantic Attention Network (SAN)}$} To train the parameter set ${\theta _{e}}$, the base approach DEM \cite{IEEEcvpr17:Zhang} proposes to minimize the differences between each sample visual feature and the visual embedding of the corresponding class semantic prototype. However, such a method is trained with stochastic gradient descent (SGD) in a batch-based fashion, which easily suffers from the class imbalance issue. In this work, we propose to train ${\theta _{e}}$ by minimizing the differences between the class semantic embeddings in the visual space and the class visual prototypes. The class visual prototypes are obtained with the same number of instances from each training class via a semantic attention network (SAN), such that each class is ensured to contribute equally for the learned model during each iteration. A simple method to get the class visual prototype $\mathbf{\tilde x}_i$ is to average the visual features of $n$ randomly selected samples directly for $i$-th class (denoted in green line in Fig. \ref{fig2}), that is:
\begin{equation}
\mathbf{\tilde x}_i^a = \frac{1}{n}\sum\nolimits_{j = 1}^n {{{\bf{x}}_j}} ,
\end{equation}
where $\mathbf{\tilde x}_i^a$ denotes one kind of visual prototype via directly averaging the selected sample feature.

As illustrated in Fig. \ref{fig1}, the class visual prototype via directly averaging the selected sample features is considerably affected by the outliers such as the images with noisy background or non-typical characteristics, which will cause the prediction model deviates from the fact. To this end, we fuse the selected instances to obtain the semantics-guided class visual prototype by distinguishing the selected instances based on their relevances to their corresponding class semantic prototypes (denoted in blue line in Fig. \ref{fig2}). Specifically, we first compute the cosine similarity $s_{ij}$ between the embedded class semantic prototype and each single selected visual feature, i.e.
\begin{equation}
{s_{ij}} = \cos ({{\bf{\hat x}}_i},{\bf{x}}_i^j) = \frac{{{{{\bf{\hat x}}}_i}^T{\bf{x}}_i^j}}{{||{{{\bf{\hat x}}}_i}||||{\bf{x}}_i^j||}}, ~j \in [1,n],
\end{equation}
where ${\mathbf{x}}_i^j$ is the visual feature of the $j$-th selected sample from $i$-th class.

Then we obtain the attention weights $\alpha_{ij}$ for each single sample via a $softmax$ function:
\begin{equation}
{\alpha _{ij}} = \frac{{exp({s_{ij}})}}{{\sum\nolimits_{t = 1}^n {exp({s_{it}})} }},~j \in [1,n].
\end{equation}

To attend on each selected sample with respect to the class semantic prototype, we define a weighted combination for each original visual feature to get the semantics-guided class visual prototype $\mathbf{\tilde x}_i^b$:
\begin{equation}
{\bf{\tilde x}}_i^b = \sum\nolimits_{j = 1}^n {{\alpha _{ij}}{\bf{x}}_i^j}.
\end{equation}

The averaged class visual prototype $\mathbf{\tilde x}_i^a$ encodes the general feature of the selected samples, which preserves the global information. In contrast, the semantics-guided class visual prototype $\mathbf{\tilde x}_i^b$ focuses on the representativeness of each single sample guided by the class semantic prototype, which reflects the special information provided by the individual. Thus, it is necessary to fuse $\mathbf{\tilde x}_i^a$ and $\mathbf{\tilde x}_i^b$ to get a more robust class visual prototype (denoted in red line in Fig. \ref{fig2}). On one side, the noisy information will be suppressed in the fused prototype. On the other side, the individual information in $\mathbf{\tilde x}_i^b$ serves as a supplement for the global information to reduce the information loss caused by the mean operation. The formulation of the fused class visual prototype $\mathbf{\tilde x}_i^c$ is expressed as follows:
\begin{equation}
{\bf{\tilde x}}_i^c = {\lambda _p}{\bf{\tilde x}}_i^a + (1 - {\lambda _p}){\bf{\tilde x}}_i^b,
\end{equation}
where $\lambda_p$ is the tunable parameter to balance $\mathbf{\tilde x}_i^a$ and $\mathbf{\tilde x}_i^b$.

\subsection{Loss Functions}
To align the common semantic information between the visual space and the semantic space, the embedded class semantic prototype is forced to be close to the corresponding visual prototype, which is formulated as follows:
\begin{equation}
{L_1} = \arg \mathop {\min }\limits_{{\theta _e}}  {\left\| {{{{\bf{\hat x}}}_i} - {{{\bf{\tilde x}}}_i}} \right\|_2^2} ,~{{\bf{\tilde x}}_i} \in \left\{ {{\bf{\tilde x}}_i^a,{\bf{\tilde x}}_i^b,{\bf{\tilde x}}_i^c} \right\},
\end{equation}
where ${\left\|  \cdot  \right\|_2}$ denotes the $l_2$ norm.

As shown in Fig. \ref{fig2}, to further model $\mathbf{\hat x}_i$ and $\mathbf{\tilde x}_j$ into a similar structure, we embed both $\mathbf{\hat x}_i$ and $\mathbf{\tilde x}_j$ to a latent space with the same hidden dimensionality as that in SEN via a function $g(\cdot)$, where cosine similarity is applied to ensure the matching relevance between the two modality prototypes to be maximized. Note that the parameters are shared during the projection, which are represented as $\theta_{s}$. Then we obtain the loss function $L_2$:
\begin{equation}
{L_2} = \arg \mathop {\max }\limits_{{\theta _s}} \frac{{{g_{{\theta _s}}}{{({{{\bf{\hat x}}}_i})}^T}{g_{{\theta _s}}}({{{\bf{\tilde x}}}_i})}}{{\left\| {{g_{{\theta _s}}}({{{\bf{\hat x}}}_i})} \right\|_2^2\left\| {{g_{{\theta _s}}}({{{\bf{\tilde x}}}_i})} \right\|_2^2}},~{{\bf{\tilde x}}_i} \in \left\{ {{\bf{\tilde x}}_i^a,{\bf{\tilde x}}_i^b,{\bf{\tilde x}}_i^c} \right\},
\end{equation}
where ${g_{{\theta _{s}}}}( \cdot )$ is the common embedding function for both $\mathbf{\hat x}_i$ and $\mathbf{\tilde x}_j$.

To further constrain the distribution of the embedded semantic prototypes and the visual prototypes for all categories, we employ a hinge loss to make ${\mathbf{\hat x}}_i$ get closer to $\mathbf{\tilde x}_j$ if $i = j$. Otherwise, ${\mathbf{\hat x}}_i$ will be pulled far away from $\mathbf{\tilde x}_j$ within a margin. To this end, we compute:
\begin{equation}
{L_3} = \mathop {\min }\limits_{{\theta _{e}}} [{\lambda _y}\left\| {{{{\bf{\hat x}}}_i} - {{{\bf{\tilde x}}}_j}} \right\|_2^2 + (1 - {\lambda _y})max[\gamma  - \left\| {{{{\bf{\hat x}}}_i} - {{{\bf{\tilde x}}}_j}} \right\|_2^2,0]],
\end{equation}
where $\gamma$ is the margin controlling the separating degree for the visual and semantic prototypes from different classes and $\lambda_y$ is the 0-1 similarity between $\mathbf{\hat x}_i$ and $\mathbf{\tilde x}_j$, i.e., ${\lambda _y} = 1$, if $ i$ = $j$; otherwize, ${\lambda _y} = 0.$

The final loss function is defined as follows:
\begin{equation}
L = {\lambda _q}{L_1} + (1 - {\lambda _q}){L_3} - \beta {L_2} + \left\| {{\theta _{e}}} \right\|_2^2 + \left\| {{\theta _{s}}} \right\|_2^2,
\end{equation}
where $\lambda_q$ is the weight parameter to tune the loss $L_1$ and the loss $L_3$, and $\beta$ is the hyper-parameter to match the relevant visual and semantic prototypes from the same class.

\subsection{Apply to ZSC}
During the training stage, the model is well trained to balance the generalization ability for all classes via constructing the class visual prototypes. In the test process, given a test instance ${{\bf{x}}_u} \in {{\bf{X}}_U}$ and a set of test semantic features ${{\bf{A}}_U} = \{ {{\bf{a}}_1},...,{{\bf{a}}_i},...,{{\bf{a}}_t}\}  \in {\mathbb{R}^{t \times q}}$ from $t$ candidate classes, ZSC is achieved in two steps. First, the test semantic features set ${{\bf{A}}_U}$ is embedded into the visual space to obtain the test class prototypes set ${{\bf{\hat X}}_U}$. After that, the prediction of ${{\bf{x}}_u}$ is performed by minimizing its distance $d$ to ${{\bf{\hat X}}_U}$ in the visual space:
\begin{equation}
{y_u^*} = \arg \mathop {\min }\limits_{{y_u}}d({{\bf{x}}_u},{{\bf{\hat X}}_U}),~{y_u} \in {Y_U}.
\end{equation}

\section{Experiments}
\begin{figure}[t]
	\centering
	\includegraphics[scale=0.65]{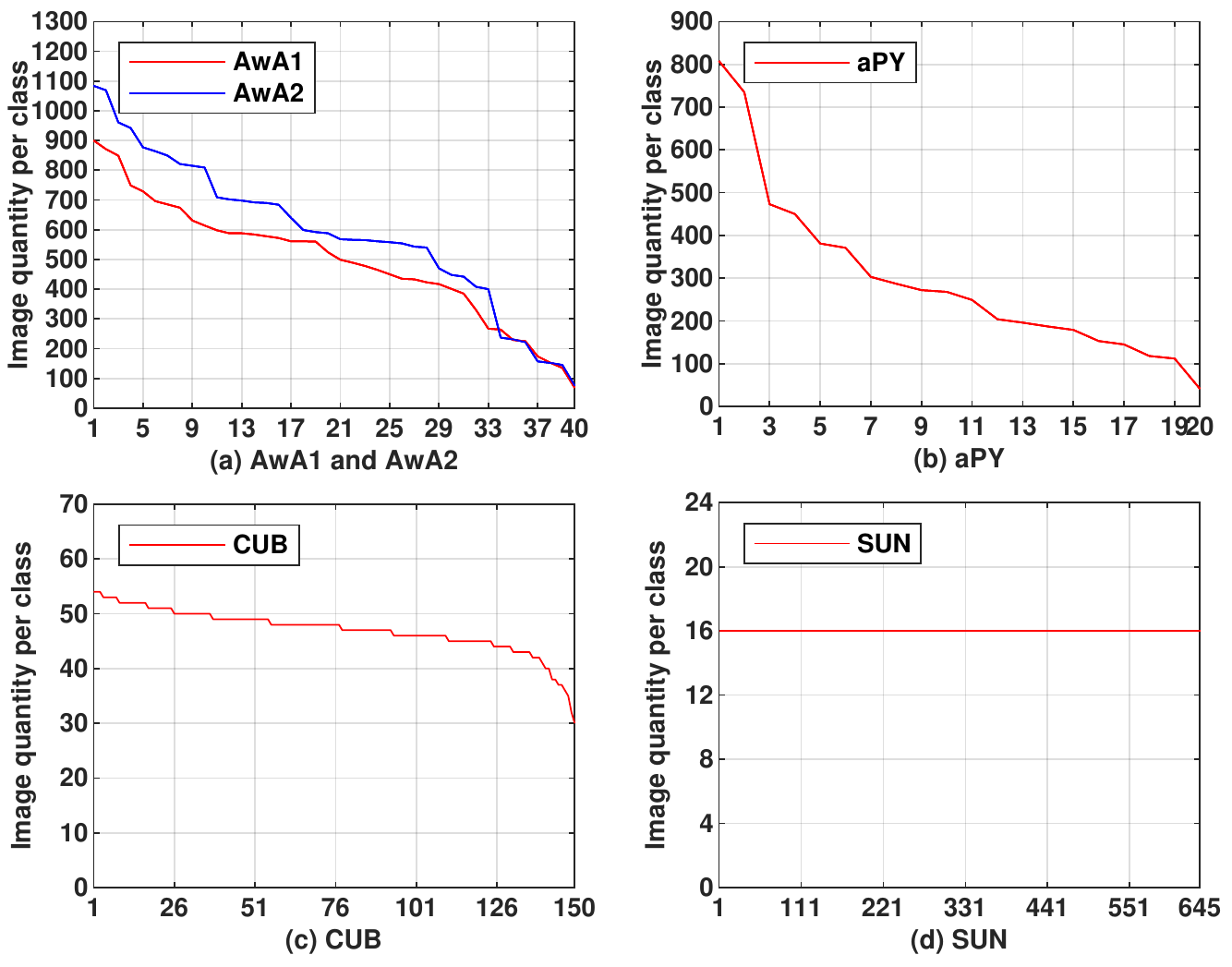}
	\caption{The statistics of the image numbers per class for AwA1, Aw2, aPY, CUB, and SUN datasets.}
	\label{fig3}
\end{figure}

	In this section, we perform experiments on the benchmark datasets of ZSC. Note that our proposed method focuses on the class imbalance scenario. As shown in Fig. \ref{fig3}, for the five ZSC benchmark datasets, i.e., AwA1, AwA2, aPY, CUB, and SUN, we count the numbers of images per class in each dataset and sort them in a descending order. It can be observed that the image number in AwA1, AwA2, and aPY datasets generally follows a long-tailed distribution, which indicates that the classes with a small number of samples have lower probability to be selected during the traditional batch-based training fashion. In contrast, the CUB dataset changes smoothly and the number in SUN dataset is fixed for all training classes. Thus the class imbalance issue does not happen for both CUB and SUN datasets. Besides, we also compute the standard deviation based on the instance number per class for the five datasets, which is 202, 255, 197, 4 and 0, respectively, further confirming the differences of image numbers in different datasets. To this end, we reasonably select AwA1, AwA2, and aPY datasets in this work. Note that we also conduct experiments on the balanced datasets CUB and SUN and obtain fairly acceptable performances. To validate the effectiveness of our method on class imbalance issue, we emphatically analyze the experimental results on the three imbalanced datasets, i.e., AwA1, AwA2, and aPY dataset in the following.

\subsection{Datasets and Settings}
\paragraph*{\textbf{Datasets}} Animals with Attributes (\textbf{AwA1}) \cite{IEEEpami:Lampert} dataset contains 30,475 images from 50 classes. Each class is annotated with an 85-dimenasional attribute vector. Animals with Attributes2 (\textbf{AwA2}) \cite{IEEEcvpr17:Xian} dataset collects 37,222 publicly available images sharing the same 50 classes and class semantic attributes with AwA1. Note that the images in AwA2 do not overlap with images in AwA1. Attribute Pascal and Yahoo (\textbf{aPY}) \cite{IEEEcvpr09:Farhadi} is a dataset with 15,399 images from 32 classes in total, among which the 20 classes from Pascal are used for training and the 12 classes from Yahoo for testing. The dataset provides 64 attributes for each class. For a fair comparison, we follow the new proposed PS seen/unseen split in \cite{IEEEcvpr17:Xian} and use the ResNet deep features and attributes released in \cite{IEEEcvpr17:Xian} as the visual and semantic representation, respectively.
\paragraph*{\textbf{Evaluation Metrics}} We conduct experiments on the settings of both TZSC and GZSC. For the TZSC, we use the top-1 accuracy per class (\textbf{T}) to measure the performance. For the GZSC, the search space is extended to all classes. We follow the harmonic mean (\textbf{H}) proposed in \cite{IEEEcvpr17:Xian} for evaluation, which is computed as:
\begin{equation}
{\bf{H}} = \frac{{2 * {\bf{u}} * {\bf{s}}}}{{{\bf{u}} + {\bf{s}}}},
\end{equation}
where \textbf{u} and \textbf{s} denote the prediction accuracy in the space spanned by all classes for the images from seen classes and unseen classes, respectively.

\paragraph*{\textbf{Settings}} (1) Neural Network. The SEN module is designed as a three-layer neural network. The neurons of the hidden layer in SEN are set to 1,000, 1,600, and 1,024 for the three datasets of AwA1, AwA2 and aPY, respectively. We apply ReLU activation function for full connected layers for all datasets.

(2) Hyper-parameters. The margin $\gamma$ is set to 2, 1, and 0.5 for AwA1, AwA2 and aPY, respectively. The class visual prototype balance parameter $\lambda_p$ is set to 0.4, 0.3, and 0.5, respectively. The loss balance parameter $\lambda_q$ is set to 0.4, 0.4, and 0.2, respectively. The number of selected samples $n$ during each iteration is set to 10, 20, and 30, respectively.

%{\color{red}For simplicity, we refer our proposed method as AFEM in the following sections.}

%\renewcommand\arraystretch{1.3}
\begin{table}[t]
	\caption{Comparison results on TZSC (in \%). The best performance is marked in bold.}
	\label{Table 1}
	\centering \begin{tabular}{|p{2.3cm}|p{1cm}<{\centering}p{1cm}<{\centering}p{1cm}<{\centering}|p{1cm}<{\centering}|}
		\hline
		Method             &AwA1  &AwA2  &aPY   &Average \\
		\hline
		\hline
		DAP \cite{IEEEpami:Lampert}               &44.1  &46.1  &33.8  &41.3 \\
		IAP \cite{IEEEpami:Lampert}               &35.9  &35.9  &36.6  &36.1 \\
		ConSE \cite{ICLR14:Norouzi}             &45.6  &44.5  &26.9  &39.0 \\
		CMT \cite{IEEEnips13:Socher}     			&39.5  &37.9  &28.0  &35.1 \\
		DeViSE \cite{IEEEnips13:Frome}  			&54.2  &59.7  &39.8  &51.2 \\
		SJE \cite{IEEEcvpr15:Akata}     			&65.6  &61.9  &32.9  &53.5 \\
		ESZSL \cite{IEEEnips15:Romera-Paredes}   			&58.2  &58.6  &38.3  &51.7 \\
		SSE \cite{iccv15:Zhang}     			&60.1  &61.0  &34.0  &51.7 \\
		SynC \cite{IEEEcvpr16:Changpinyo}   			&54.0  &46.6  &23.9  &41.5 \\
		LATEM \cite{IEEEcvpr16:Xian}   			&55.1  &55.8  &35.2  &48.7 \\
		ALE \cite{akata2015label}     			&59.9  &62.5  &39.7  &54.0 \\
		SAE \cite{IEEEcvpr17:Kodirov}     			&53.0  &54.1  &8.3   &38.5 \\
		%GFZSL \citep{verma2017simple} &68.3 &1.8 &80.3 &3.5  &63.8 &2.5 &80.1 &4.8    &49.2 &- &\textbf{-} &-  \\
		
		CDL \cite{jiang2018learning}     	&39.5  &-     &28.0     &- \\
 		%3ME[]		&60.2  &-     &-     &- \\
		Relation Net \cite{IEEEcvpr18:Yang} 		&68.2  &64.2  &-     &- \\
		SP-AEN \cite{chen2018zero}			&58.5  &-     &24.1  &- \\
		%Gaussian-Ort[]      &70.1  &70.5  &-     &- \\
		f-CLSWGAN[]	&68.2 &-     &-     &- \\
		%RAS-cGAN[] 			&67.4  &-     &-     &- \\
		TVN-deep \cite{zhang2019triple} &68.8 &- &\textbf{41.3} &-\\
		GAZSL \cite{IEEEcvpr18:Zhu}  	&68.2  &69.0  &41.1     &59.4 \\
		DEM \cite{IEEEcvpr17:Zhang}     	&68.4  &67.1  &35.0  &56.8 \\
		\textbf{SCILM (Ours)} &\textbf{72.5}  &\textbf{70.1}  &38.4  &\textbf{60.3} \\
		\hline
	\end{tabular}
\end{table}

\subsection{Comparison with the State-of-the-art}
To demonstrate the effectiveness of our proposed SCILM, we evaluate our proposed SCILM against twenty popular models on both TZSC and GZSC settings. These approaches mainly fall into two groups. (1) Embedding-based approaches. DAP \cite{IEEEpami:Lampert}, IAP \cite{IEEEpami:Lampert}, ConSE \cite{ICLR14:Norouzi}, CMT \cite{IEEEnips13:Socher}, DeViSE \cite{IEEEnips13:Frome}, SJE \cite{IEEEcvpr15:Akata}, ESZSL \cite{IEEEnips15:Romera-Paredes}, SSE \cite{iccv15:Zhang}, SynC \cite{IEEEcvpr16:Changpinyo}, LATEM \cite{IEEEcvpr16:Xian}, ALE \cite{akata2015label}, SAE \cite{IEEEcvpr17:Kodirov}, CDL \cite{jiang2018learning}, Relation Net \cite{IEEEcvpr18:Yang} and DEM \cite{IEEEcvpr17:Zhang} are modeled by exploring a semantic embedding, visual embedding or common embedding space. (2) Generative adversarial based approaches. SP-AEN \cite{chen2018zero}, f-CLSWGAN \cite{xian2018feature}, COSMO+FCLSWGAN \cite{atzmon2019adaptive}, and GAZSL \cite{IEEEcvpr18:Zhu} combine the training measure of GAN \cite{goodfellow2014generative} with ZSC via different regularizations or model structures. Our SCILM belongs to the former group. SCILM follows the same embedding direction from the semantic space to the visual space and adopts the similar ridge regression model structure with DEM \cite{IEEEcvpr17:Zhang}. Thus, DEM \cite{IEEEcvpr17:Zhang} is adopted as the baseline method in our work. Note that our SCILM focuses on the class imbalance issue that is neglected in DEM \cite{IEEEcvpr17:Zhang} and is trained in a novel sample-balanced manner instead of the traditional batch-based training style adopted in DEM \cite{IEEEcvpr17:Zhang}. 

For the performances of the TZSC task, as shown in Table \ref{Table 1}, SCILM achieves the best performance on AwA1, AwA2 datasets, and the Average metric, which outperforms the runners up in 4.1\%, 1.1\% and 0.9\%, respectively. As for the aPY dataset, SCILM is also comparable to the best approach GAZSL. Besides, SCILM beats all the embedding based methods. Compared with the baseline method DEM \cite{IEEEcvpr17:Zhang}, SCILM significantly improves the performance by 4.1\%, 3.0\%, 3.4\%, and 3.5\%, respectively. Note that compared with the generative adversarial based approaches, SCILM has a much simpler network structure and does not require complicated adversarial training process, which is more stable and efficient.

\begin{table*}[t]
	\caption{Comparison results on GZSC (in \%). The best performance is marked in bold.}
	\label{Table 2}
	\centering
	\begin{tabular}{|p{3.3cm}|p{1cm}<{\centering}p{1cm}<{\centering}p{1cm}<{\centering}|p{1cm}<{\centering}p{1cm}<{\centering}p{1cm}<{\centering}|p{1cm}<{\centering}p{1cm}<{\centering}p{1cm}<{\centering}|}
		\hline
		\multirow{2}{*}{Method}&\multicolumn{3}{c|}{\textbf{AwA1}}  &\multicolumn{3}{c|}{\textbf{AwA2}}   &\multicolumn{3}{c|}{\textbf{aPY}}\\
		%\cline{2-10}
		&\textbf{u} & \textbf{s} &\textbf{H}   &\textbf{u} &\textbf{s} &\textbf{H}  &\textbf{u} & \textbf{s} &\textbf{H}\\
		\hline
		\hline
		DAP \cite{IEEEpami:Lampert}     &0.0 &88.7 &0.0             &0.0 &84.7 &0.0              &4.8 &78.3  &9.0 \\
		IAP \cite{IEEEpami:Lampert}     &2.1 &78.2 &4.1             &0.9 &87.6 &1.8              &5.7  &65.6 &10.4 \\
		ConSE \cite{ICLR14:Norouzi}   &0.4 &88.6 &0.8             &0.5 &90.6 &1.0              &0.0  &91.2 &0.0 \\
		CMT \cite{IEEEnips13:Socher}     &8.4 &86.9 &15.3            &8.7 &89.0 &15.9             &10.9 &74.2 &19.0  \\
		DeViSE \cite{IEEEnips13:Frome}  &13.4 &68.7 &22.4  		  &17.1 &74.7 &27.8            &4.9  &76.9 &9.2 \\
		SJE \cite{IEEEcvpr15:Akata}      &11.3 &74.6 &19.6           &8.0 &73.9 &14.4             &3.7 &55.7 &6.9\\
		ESZSL \cite{IEEEnips15:Romera-Paredes}   &6.6 &75.6  &12.1           &5.9 &77.8 &11.0             &2.4 &70.1 &4.6 \\
		SSE \cite{iccv15:Zhang}     &7.0 &80.5 &12.9            &8.1 &82.5 &14.8             &0.2 &78.9 &0.4 \\
		SynC \cite{IEEEcvpr16:Changpinyo}    &8.9 &87.3 &16.2            &10.0 &90.5 &18.0            &7.4 &66.3 &13.3 \\
		LATEM \cite{IEEEcvpr16:Xian}   &7.3 &71.7 &13.3            &11.5 &77.3 &20.0            &0.1 &73.0 &0.2 \\
		ALE \cite{akata2015label}     &16.8 &76.1 &27.5           &14.0 &81.8 &23.9            &4.6 &73.7 &8.7\\
		SAE \cite{IEEEcvpr17:Kodirov}     &1.8 &77.1 &3.5             &1.1 &82.2 &2.2              &0.4 &80.9 &0.9  \\
		%GFZSL \citep{verma2017simple} &68.3 &1.8 &80.3 &3.5  &63.8 &2.5 &80.1 &4.8    &49.2 &- &\textbf{-} &-  \\
		
		CDL \cite{jiang2018learning}       &28.1 &73.5 &40.6           & - & - & -                 &19.8 &48.6 &28.1\\
		%3ME[2019arxiv] &55.5 &65.7 &60.2		 & - & - & - 				 & - & - & - \\
		Relation Net \cite{IEEEcvpr18:Yang}  &31.4 &\textbf{91.3} &46.7        &30.0 &\textbf{93.4} &45.3	 & - & - & - \\
		SP-AEN \cite{chen2018zero}        &23.3 &90.9 &37.1        & - & - & -         &13.7 &63.4 &22.6\\
%		cycle-CLSWGAN \cite{felix2018multi} &\textbf{56.9} &64.0 &60.2		 & - & - & -		 &- &- &- \\
		TVN-deep \cite{zhang2019triple}  &27.0 &67.9 &38.6 &- &- &- &16.1 &66.9 &25.9 \\
		f-CLSWGAN \cite{xian2018feature}  &57.9 &61.4 &59.6          &- &- &-   	     &- &- &- \\
		COSMO+FCLSWGAN \cite{atzmon2019adaptive} &64.8 &51.7 &57.5     &- &- &-   	     &- &- &- \\
		%		RAS-cGAN[]       &38.7 &74.6 &51.0          & - &- &-                    &- &- &- \\
		GAZSL \cite{IEEEcvpr18:Zhu}  &29.6 &84.2 &43.8          &35.4 &86.9 &50.3            &14.2	&\textbf{78.6}	&24.0\\
		DEM \cite{IEEEcvpr17:Zhang}        &32.8 &84.7  &47.3          &30.5	&86.4 &45.1	         &11.1 &75.1 &19.4\\
		\textbf{SCILM (Ours)} &52.9 &72.2 &\textbf{61.1}   &\textbf{39.2} &77.3 &\textbf{52.1}			 &\textbf{22.8} &62.7 &\textbf{33.4} \\
		%PSR &- & - & - &- &63.8 & 20.7 &73.8 &32.3 & 56.0 & 24.6 &54.3 &33.9\\
		%GAZSL \citep{zhu2018generative} &68.2 &25.7 &82.0 &39.2   &69.0 &27.0 &82.4 &40.6  &55.8 &31.7 &61.3 &41.8\\
		%RELATION NET \citep{yang2018learning}&68.2 &31.4 &\textbf{91.3} &46.7   &64.2 &30.0 &\textbf{93.4} &45.3 &55.6 &38.1 &61.6 &47.0 \\
		%f-CLSWGAN [36]57:7 43:7 49:7 36:6 42:6 39:4 61:4 57:9 59:6 68:9 52:1 59:4
		\hline
	\end{tabular}
\end{table*}

From the comparison results of the GZSC task in Table~~\ref{Table 2}, we observe that SCILM beats all the runner up approaches of f-CLSWGAN \cite{xian2018feature}, GAZSL \cite{IEEEcvpr18:Zhu}, and CDL \cite{jiang2018learning} on the \textbf{H} metric by 1.5\%, 1.8\%, and 5.3\% on AwA1, AwA2, and aPY datasets, respectively. On the metric of \textbf{u}, SCILM achieves the best performance on both AwA2 and aPY datasets and gains in 3.8\% and 3.8\%, respectively. It is also competitive on AwA1 dataset and ranks third. It is clearly that the encouraging improvement of \textbf{H} metric mainly benefits from the performance promotion of \textbf{u} metric. The results further prove that SCILM effectively enhances the model generalization ability and robustness for unseen categories by proposing a semantically guided class-specific sample training model to alleviate the class imbalance issue during batch training. We also notice that SCILM is relatively unsatisfactory on \textbf{s} metric. It is quite understandable for that the generalization ability for seen and unseen classes is a trade-off.
\begin{table}[!t]
	\caption{The evaluation of different visual prototypes on \textbf{H} metric under GZSC setting (in \%). The best performance is marked in bold.}
	\label{Table 3}
	\centering \begin{tabular}{|p{2.5cm}|p{1cm}<{\centering}p{1cm}<{\centering}p{1cm}<{\centering}|p{1cm}<{\centering}|}
		\hline
		Method &AwA1  &AwA2  &aPY   &Average \\
		\hline
		\hline
		DEM (baseline) \cite{IEEEcvpr17:Zhang}		 &47.3  &45.1  &19.4  &37.3 \\
		SCILM-a               &51.8  &44.1  &10.1  &35.3 \\
		SCILM-b               &53.9  &46.8  &23.2  &41.3 \\
		SCILM-c               &\textbf{61.1}  &\textbf{52.1}  &\textbf{33.4}  &\textbf{48.9} \\
		\hline
	\end{tabular}
\end{table}
\subsection{Ablation Studies}
In order to explore the classification performance of different class visual prototypes and validate the effectiveness of the SAN module, we conduct ablation experiments on the three datasets. Table \ref{Table 3} summarizes the results, for which SCILM-a, SCILM-b, and SCILM-c denote the proposed model constructed with the directly averaged visual prototype $\mathbf{\tilde x}_i^a$, the semantics-guided class visual prototype $\mathbf{\tilde x}_i^b$ and the fused class visual prototype $\mathbf{\tilde x}_i^c$, respectively. We have the following observations from Table \ref{Table 3}.

(1) It is obvious that SCILM-c dramatically increases SCILM-b by 7.2\%, 5.3\%, 10.2\%, and 7.6\% on AwA1, AwA2, aPY datasets, and the Average metric, respectively. The improvements further confirm that combining the general mean feature with the semantically related individual feature tends to produce more robust visual representation thus promoting the model performance.

(2) SCILM-a hardly outperforms the baseline method DEM \cite{IEEEcvpr17:Zhang}. Compared with DEM \cite{IEEEcvpr17:Zhang}, SCILM-a is achieved by directly averaging the visual features of the randomly selected samples from each class as the class visual prototype, while DEM \cite{IEEEcvpr17:Zhang} is trained with the data batch randomly selected from all training data. We argue the reasons lie in that directly employing the mean feature of the selected samples may result in the loss of feature diversity for each category. Despite that the prediction performance for the classes with small number of samples will be improved for considering the class imbalance issue, the performance for the classes with sufficient samples may be degraded to a certain extent.

(3) The visual prototype $\mathbf{\tilde x}_i^b$ guided by SAN component achieves obvious improvements compared with $\mathbf{\tilde x}_i^a$ on three datasets. It validates that the semantics-guided visual prototype considering the contributions of each selected sample is able to preserve the semantic relations and suppress the influence of the image outliers in the visual space, thereby making the class visual prototypes be more discriminative.

\subsection{Discussions on the Effectiveness for Small Sample Classes}
\begin{figure*}[t]
	\centering
	\includegraphics[scale=0.88]{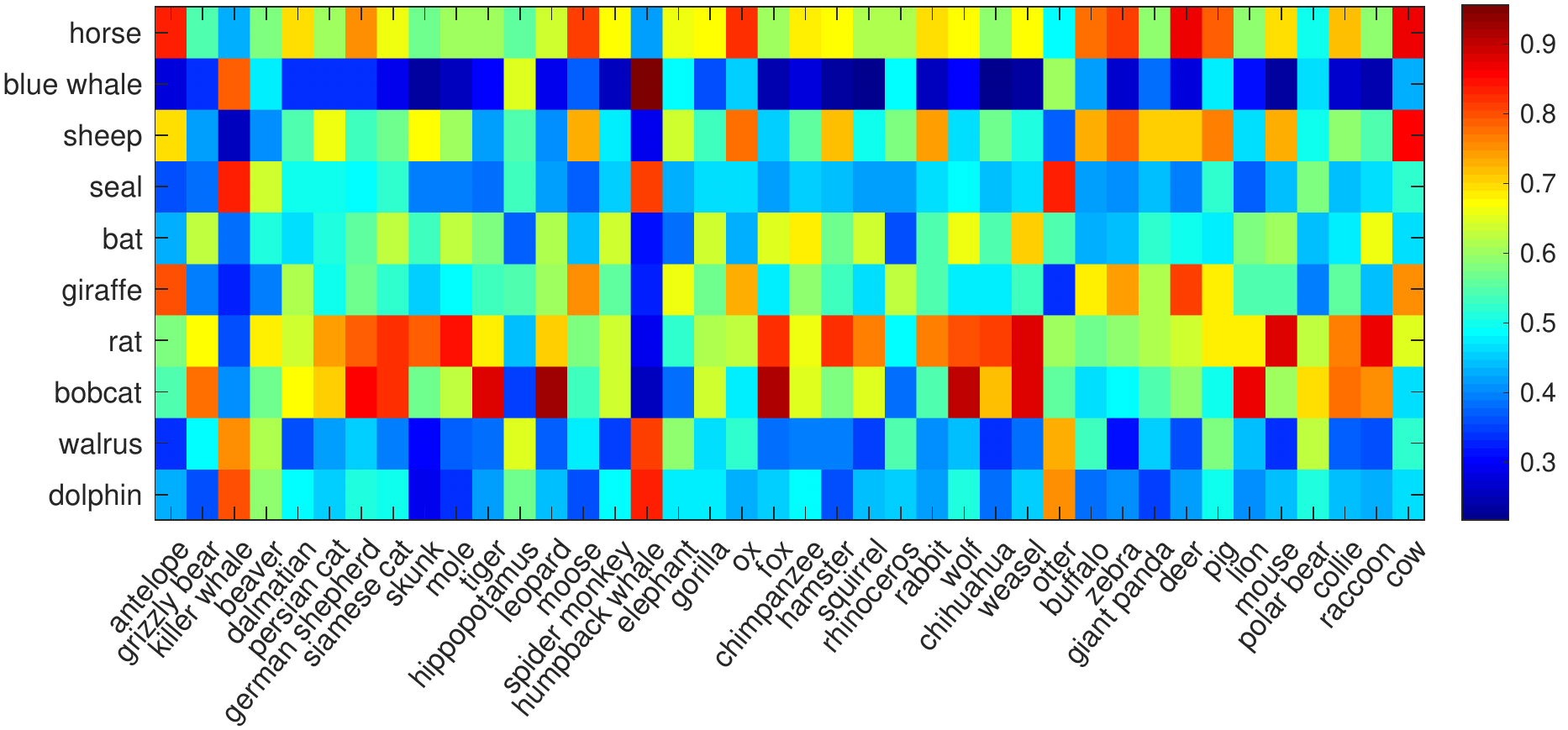}
	\caption{The cosine semantic similarity matrix of the unseen and seen classes for AwA1 dataset. The color bar on the right indicates the similarity scores.}
	\label{fig4}
\end{figure*}

\begin{figure}[t]
	
	\centering
	\includegraphics[scale=0.75]{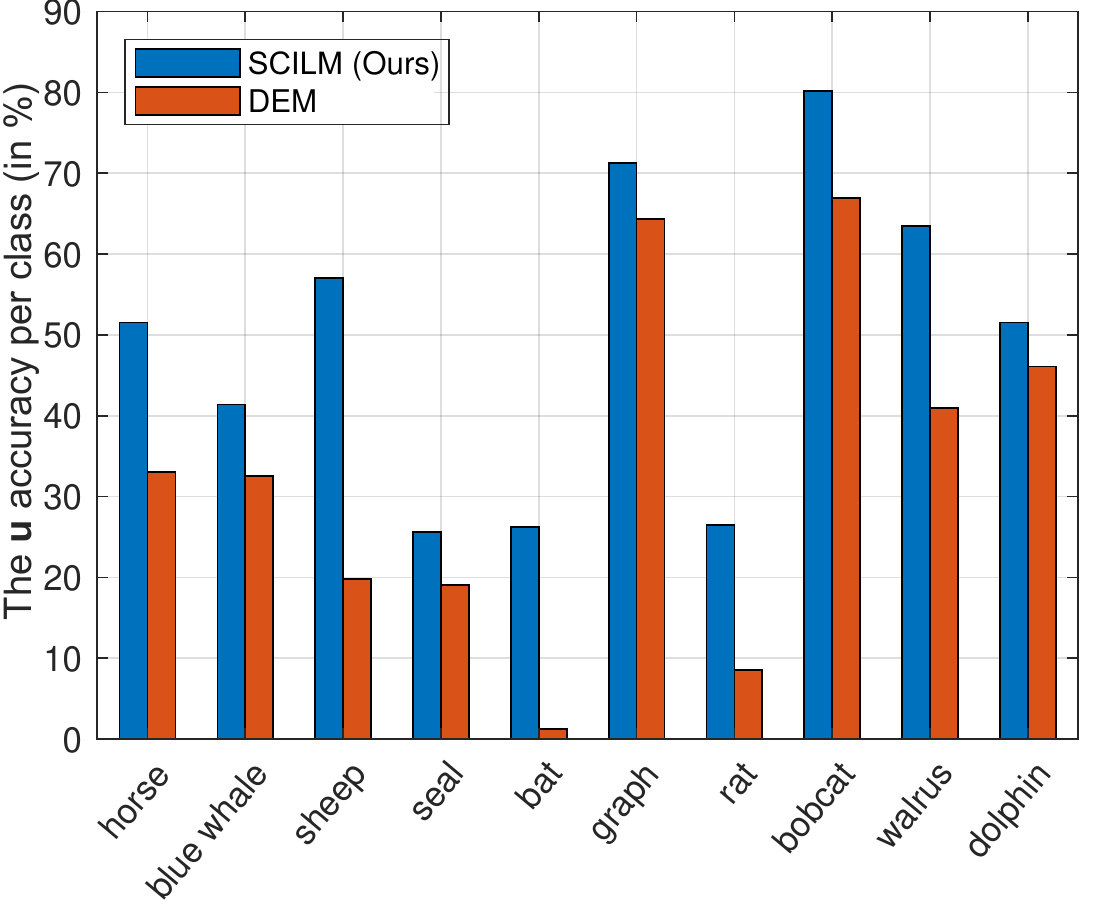}
	\caption{\upshape The \textbf{u} performance for each unseen class of both SCILM and DEM \cite{IEEEcvpr17:Zhang} on AwA1 dataset under the GZSC setting.}
	\label{fig5}
\end{figure}

We infer that the prediction performance for the unseen classes that are closely related to some sample-scarce seen classes is improved via our SCILM training mode. In this subsection, we conduct additional experiments to further estimate the impacts of SCILM on the unseen classes that are only related to sample-scarce seen classes. To quantitatively measure the relatedness between the unseen and the seen classes, we calculate the cosine similarity between each pair of them, as presented in Fig. \ref{fig4}. The \textbf{u} metric under GZSC setting is selected to measure the model's generalization ability for unseen classes. We take AwA1 dataset as example and present the classification results of SCILM and the baseline method DEM \cite{IEEEcvpr17:Zhang} for ten unseen classes in Fig. \ref{fig5}.

\begin{figure*}[t]
%\begin{center}
	\centering
	\includegraphics[scale=0.75]{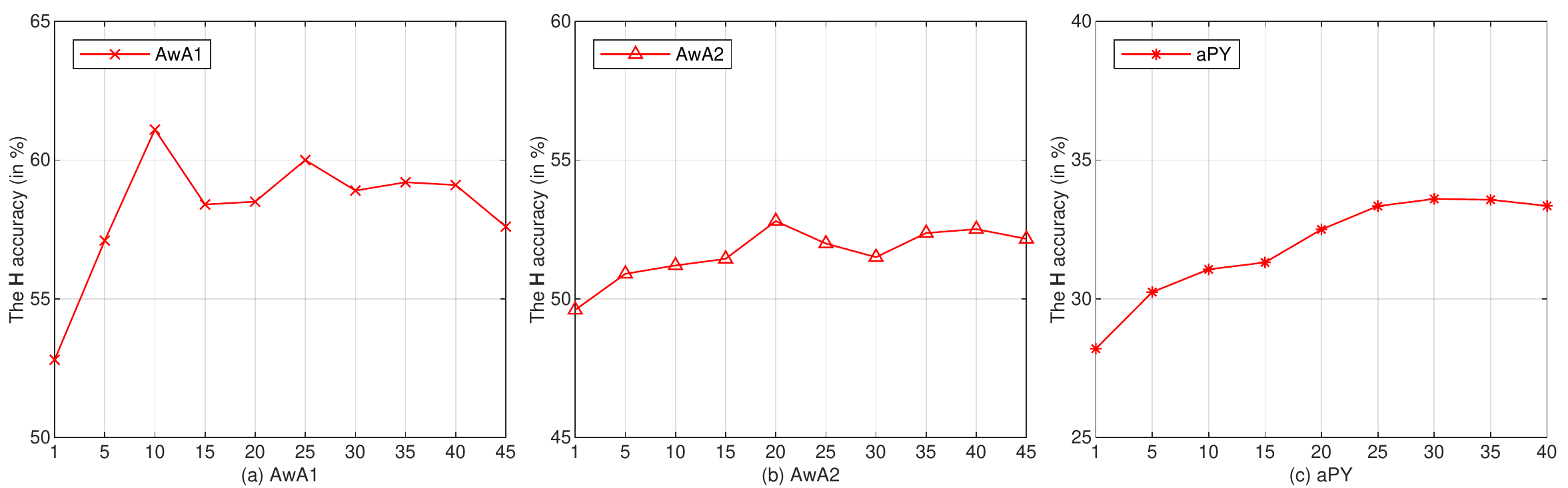}
	\caption{The accuracy under \textbf{H} metric obtained by our SCILM respectively to the number of selected images for AwA1, AwA2, and aPY datasets, respectively.}
	\label{fig6}
%\end{center}
\end{figure*}

Notably, SCILM achieves significant improvements for the classes with poor performance by DEM \cite{IEEEcvpr17:Zhang} such as ``bat" and ``rat", meanwhile further promoting the well-classified classes such as ``graph" and ``bobcat" by DEM \cite{IEEEcvpr17:Zhang}. From Fig. \ref{fig4}, it can be observed that the most related seen category to ``bat" is ``weasel" and the top three related seen categories to ``rat" are ``mouse", ``weasel", and ``mole". The numbers of images for the class ``mouse", ``weasel", and ``mole" are 174, 225, and 68, respectively. According to Fig. \ref{fig3}, it is clear that all of the three categories have a small sample size, for which the traditional batch-based training fashion will reach a higher error rate in the unbalanced training set. This will result in poor knowledge transfer ability especially for the class like ``bat" which is only highly correlated with one seen category with few samples.

In contrast, SCILM considers the class imbalance issue in the training process. By adopting a class-specific random sample extraction strategy, the small sample categories are encouraged to have the same decision effect as the sample-abundant categories in each iteration. Meanwhile, the whole model is guided by semantics to distinguish the representativeness of each selected sample and generate discriminative class visual prototype, which effectively alleviates the class imbalance problems and improves the overall classification performance of the model.

\begin{figure*}[t]
	\centering
	\includegraphics[scale=0.8]{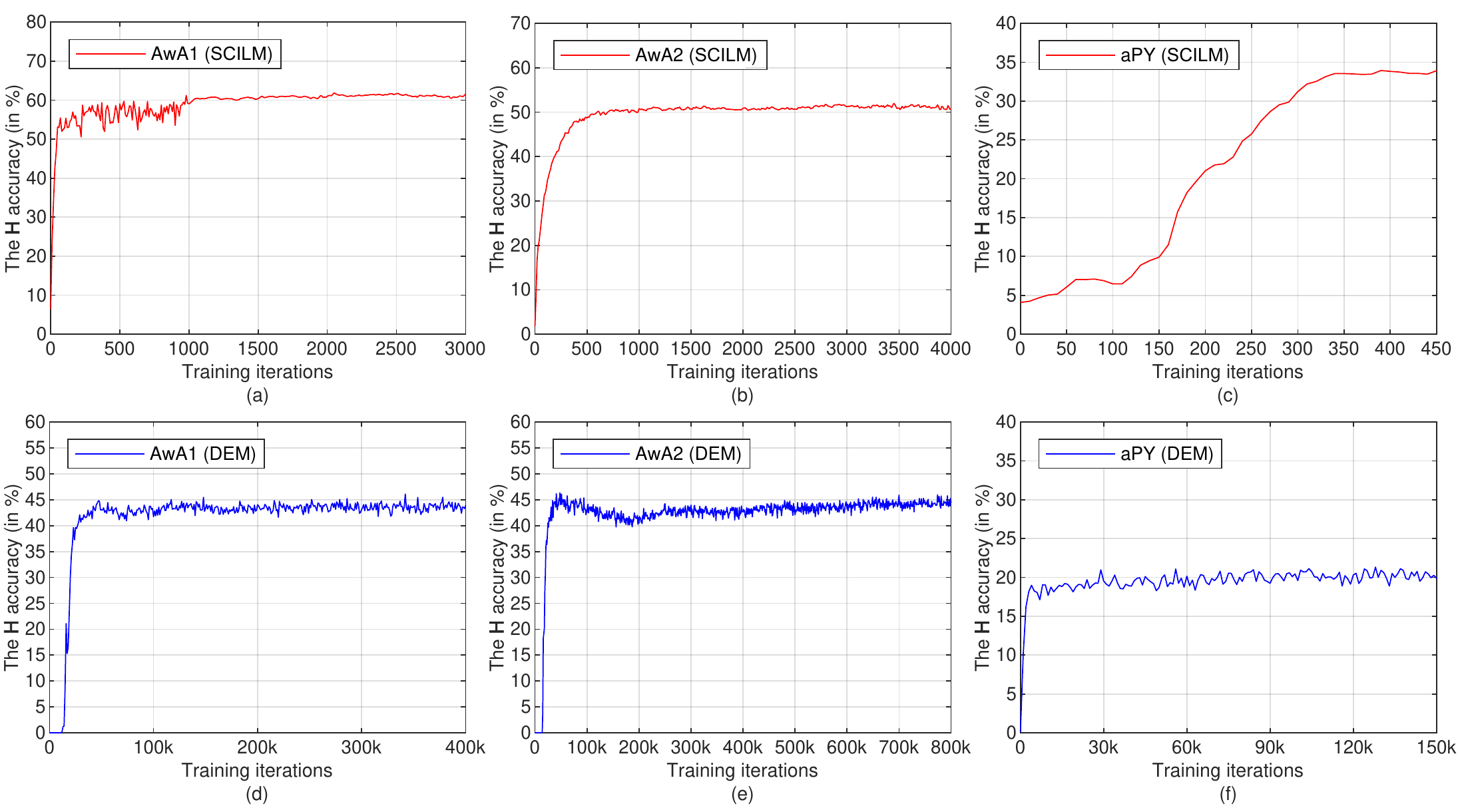}
	\caption{The efficiency evaluation for both SCILM and DEM on AwA1, AwA2, and aPY datasets under the \textbf{H} metric of GZSC setting, respectively.}
	\label{fig7}
\end{figure*}

\subsection{Analysis on the Number of the Selected Samples}
Our proposed SCILM randomly selects a certain number of samples during each training iteration to make all the training classes contribute equally to the generalized model. In this subsection, we conduct experiments to evaluate the effects of the selected sample size $n$ based on the \textbf{H} performance of GZSC setting for the three datasets. The results are presented in Fig. \ref{fig6}. First, we observe that for all three datasets, the general trend of \textbf{H} performance curve rises firstly and then declines. The number $n$ with the best \textbf{H} performance for the three datasets is 10, 20, and 30, respectively. The results indicate that using only dozens of or even a few images per class, our proposed SCILM achieves superior performance, which provides a potential solution for sample-imbalanced ZSC tasks.

Second, we notice that choosing more samples does not always bring better performance. The reasons may lie in that the difference between the mean visual prototypes tends to be smaller for each iteration when more samples are selected. Thus the scattered data distribution will be reduced, causing the loss of class information diversity. Compared with AwA1 and AwA2 datasets, aPY dataset has larger intra-class difference, thereby requiring more samples to adjust the class data distribution to reach peak performance.

\subsection{Efficiency Evaluation}
In contrast to the traditional batch-based training strategy, our proposed SCILM randomly selects a fixed number of images from each training class during each iteration to synthesize discriminative class visual prototypes, which alleviates the class imbalance issue effectively. Moreover, SCILM not only improves the accuracy but also promotes the efficiency with the novel training process and a relatively simple framework. In this subsection, we evaluate the training efficiency of SCILM and the baseline method DEM \cite{IEEEcvpr17:Zhang} on AwA1, AwA2, and aPY datasets under the \textbf{H} metric of GZSC setting. Note that our model runs on Linux Ubuntu 14.04 with an NVIDIA 11G GeForce GTX 1080 Ti GPU.

As demonstrated in Fig. \ref{fig7}, for AwA1 and AwA2 datasets, our SCILM only requires about 2,500 and 3,500 iterations respectively to achieve the convergence, which provides thousand orders of magnitude compared with DEM that requires nearly 300k and 700k iterations, respectively. As for aPY dataset with smaller quantity of images, the performance plateaus only with 400 iterations by our SCILM, but about 90k iterations by DEM \cite{IEEEcvpr17:Zhang}. Besides, we also calculate the training durations for SCILM and DEM \cite{IEEEcvpr17:Zhang}. For SCILM, the totally time cost is 26.1, 40.4, and 4.1 minutes for AwA1, AwA2, and aPY datasets, respectively, while that for DEM \cite{IEEEcvpr17:Zhang} is 1.8, 2.9, and 0.8 hours, respectively. The promising results definitely confirm the efficiency of our proposed training process.

\section{Conclusion}
In this paper, we have proposed a Semantics-Guided Class Imbalance Learning Model (SCILM) to cope with ZSC tasks from a new perspective of class imbalance issue. We focused on the sample-scarce training classes and put forward a novel class sample-balanced training process instead of the traditional batch-based training fashion to balance the overall generalization ability for all classes. Specifically, we randomly selected the same number of images from each class across all training classes to form a batch during each iteration to guarantee that the sample-scarce classes contributed equally with the sample-abundant classes. We further synthesized well-represented class visual prototypes guided by the semantic relevances and developed an effective fusion strategy to combine the general and typical class information with the special information provided by each selected sample. Extensive experiments on three imbalanced ZSC benchmark datasets demonstrated the compelling performance and high efficiency of our proposed SCILM. For TZSC task, our proposed SCILM obtained comparable performances against the state-of-the-art and increased the baseline method DEM \cite{IEEEcvpr17:Zhang} by 4.1\%, 3.0\%, and 3.4\%, respectively. For GZSC task, the \textbf{H} metric was improved by 1.5\%, 1.8\%, and 5.3\% compared with the state-of-the-art methods, respectively. The experimental result analyses about the small sample classes further confirmed that our SCILM significantly improved the prediction performance of the unseen classes that were closely related to the sample-scarce seen classes. Our work achieved superior performance with only dozens of images, providing a potential solution for sample-imbalanced ZSC tasks.

%The scarcity of target class instances makes it difficult to classify them correctly by using many traditional classifiers. xianxiang indicate

% or


%\appendices
%% you can choose not to have a title for an appendix
%% if you want by leaving the argument blank
%\section{}

%\section*{Acknowledgment}
%This work was supported by the National Natural Science Foundation of China under Grants 61771329 and 61472274. Yunlong Yu also acknowledges the support of China Scholarship Council. The authors are very grateful for NVIDIA's support in providing GPUs that made this work possible.

\ifCLASSOPTIONcaptionsoff
  \newpage
\fi


\begin{thebibliography}{1}
	
	\bibitem{szegedy2015going}
	C.~Szegedy, W.~Liu, Y.~Jia, \emph{et al.}, ``Going deeper with convolutions," in \emph{Proc. Comput. Vis. Pattern Recognit.}, Boston, USA, June 2015, pp.~1-9.
	
%	\bibitem{zhang2017landmark}
%	X.~Zhang, S.~Wang, Z.~Li, \emph{et al.}, ``Landmark image retrieval by jointing feature refinement and multimodal classifier learning," \emph{IEEE Trans. Cybern.}, vol. 48, no. 6, pp.~1682–1695, Jun. 2018.
	
	\bibitem{yu2018transductive}
	Y.~Yu, Z.~Ji, X.~Li, \emph{et al.}, ``Transductive zero-shot learning with a self-training dictionary approach," \emph{IEEE Trans. Cybern.}, vol.~48, no.~10, pp.~2908-2919, Jan. 2018.
	
	\bibitem{zhang2016transfer}
	X.~Zhang, Y.~Zhuang, W.~Wang, \emph{et al.}, ``Transfer boosting with synthetic instances for class imbalanced object recognition," \emph{IEEE Trans. Cybern.}, vol.~48, no.~1, pp.~357-370, Jan. 2016.
	
	
	\bibitem{lim2016evolutionary}
	P.~Lim, C.~K.~Goh and K.~C.~Tan, ``Evolutionary cluster-based synthetic oversampling ensemble (eco-ensemble) for imbalance learning," \emph{IEEE Trans. Cybern.}, vol.~47, no.~9, pp.~2850-2861, Sept. 2016.
	
%	\bibitem{majumder2016automatic}
%	A.~Majumder, L.~Behera, and V.~K.~Subramanian, ``Automatic facial expression recognition system using deep network-based data fusion," \emph{IEEE Trans. Cybern.}, vol. 48, no. 1, pp. 103–114, Jan. 2018.
	
	\bibitem{simonyan2014very}
	K.~Simonyan, and A.~Zisserman, ``Very deep convolutional networks for large-scale image recognition," in \emph{Int. Conf. on Learn. Rep.}, Banff, Canada, Apr. 2014.
	
%	\bibitem{zheng2018pedestrian}
%	Z.~Zheng, L.~Zheng, Y.~Liang, \emph{et al.} ``Pedestrian alignment network for large-scale person re-identification," \emph{IEEE Trans. Circ. Syst. Vid. Techn.}, Oct. 2018.
	
%	\bibitem{russakovsky2015imagenet}
%	O.~Russakovsky, J.~Deng, H.~Su, \emph{et al.}, ``Imagenet large scale visual recognition challenge," in \emph{Int. Jour. Comput. Vis.}, vol. 115, no. 3, pp.~211-252, 2015.
	
	\bibitem{krizhevsky2012imagenet}
	A.~Krizhevsky, I.~Sutskever and G.~E.~Hinton, ``Imagenet classification with deep convolutional neural networks," Advances in \emph{Neural Inf. Process. Syst.}, Harrahs and Harveys, USA, Dec. 2012, pp.~1097-1105.
	
	
	\bibitem{he2016deep}
	K.~He, X.~Zhang, S.~Ren, \emph{et al.}, ``Deep residual learning for image recognition," in \emph{Proc. IEEE Conf. on Comput. Vis. Pattern Recognit.}, Las Vegas, USA, June 2016, pp.~770-778.
	
	\bibitem{wu2018deep}
	L.~Wu, Y.~Wang, X.~Li, \emph{et al.} ``Deep attention-based spatially recursive networks for fine-grained visual recognition," \emph{IEEE Trans. Cybern.}, pp. 1-12, Mar. 2018.
	
	\bibitem{IEEEpami:Lampert}
	C.~H.~Lampert, H.~Nickisch and S.~Harmeling, ``Attribute-based classification for zero-shot visual object categorization," \emph{IEEE Trans. Pattern Anal. Mach. Intell.}, vol.~36, no.~3, pp.~453-465, Mar. 2014.
	
	\bibitem{guo2017zero}
	Y.~Guo, G.~Ding, J.~Han, and Y.~Gao, ``Zero-shot learning with transferred samples," \emph{IEEE Trans. Image Process.}, vol.~26, no.~7, pp.~3277–3290, Jul. 2017.
	
	\bibitem{IEEEcvpr15:Akata}
	Z.~Akata, S.~Reed, D.~Walter, \emph{et al.}, ``Evaluation of output embeddings for fine-grained image classification," in \emph{Proc. IEEE Conf. on Comput. Vis. Pattern Recognit.}, Boston, USA, June 2015, pp.~2927-2936.
	
	
	\bibitem{IEEEnips15:Romera-Paredes}
	B.~Romera-Paredes and P.~H.~S~Torr, ``An embarrassingly simple approach to zero-shot learning," in \emph{Proc. Int. Conf. Mach. Learn.}, Lille, France, July 2015, pp.~2152-2161.
	
	\bibitem{ICLR14:Norouzi}
	M.~Norouzi, T.~Mikolov, S.~Bengio, \emph{et al.}, ``Zero-Shot Learning by Convex Combination of Semantic Embeddings," \emph{Int. Conf. on Learn. Repr.}, Banff, Canada, Apr. 2014, pp.~1-9.
	
	\bibitem{IEEEcvpr09:Farhadi}
	A.~Farhadi, I.~Endres, D.~Hoiem, \emph{et al.}, ``Describing objects by their attributes," in \emph{Proc. IEEE Conf. on Comput. Vis. Pattern Recognit.}, Miami, USA, June 2009, pp.~ 1778-1785.
	
	\bibitem{morgado2017semantically}
	P.~Morgado and N.~Vasconcelos, ``Semantically consistent regularization for zero-shot recognition," in \emph{Proc. IEEE Conf. on Comput. Vis. Pattern Recognit.}, Honolulu, Hawaii, July 2017, pp.~6060-6069.
	
	\bibitem{IEEEnips13:Mikolov}
	T.~Mikolov, I.~Sutskever, K.~Chen, \emph{et al.}, ``Distributed representations of words and phrases and their compositionality," Advances in \emph{Neural Inf. Process. Syst.}, Nevada, USA, Dec. 2013, pp.~3111-3119.
	
	\bibitem{lei2015predicting}
	J.~Lei~Ba, K.~Swersky, S.~Fidler, \emph{et al.}, ``Predicting deep zero-shot convolutional neural networks using textual descriptions," in \emph{Proc. IEEE Int. Conf. on Comput. Vis.}, Santiago, Chile, Dec. 2015, pp.~4247-4255.
	
	\bibitem{qiao2016less}
	R.~Qiao, L.~Liu, C.~Shen, \emph{et al.}, ``Less is more: zero-shot learning from online textual documents with noise suppression," in \emph{Proc. IEEE Conf. on Comput. Vis. Pattern Recognit.}, Las Vegas,
	USA, June 2016, pp.~2249-2257.
	
	\bibitem{elhoseiny2017link}
	M.~Elhoseiny, Y.~Zhu, H.~Zhang, \emph{et al.}, ``Link the head to the" beak": Zero shot learning from noisy text description at part precision," in \emph{Proc. IEEE Conf. on Comput. Vis. Pattern Recognit.}, Honolulu, Hawaii, July 2017, pp.~6299-6297.
	
	\bibitem{zhu2018generative}
	Y.~Zhu, M.~Elhoseiny, L.~Mohamed, \emph{et al.}, ``A generative adversarial approach for zero-shot learning from noisy texts," in \emph{Proc. IEEE Conf. on Comput. Vis. Pattern Recognit.}, Salt Lake, USA, June 2018, pp.~1004-1013.
	
	\bibitem{Technical11:Wah}
	C.~Wah, S.~Branson, P.~Welinder, \emph{et al.}, ``The Caltech-UCSD Birds-200-2011 Dataset", \emph{Technical rep.}, 2011.
	
	\bibitem{IEEEcvpr17:Xian}
	Y.~Xian, B.~Schiele, and Z.~Akata, ``Zero-shot learning-The Good, the Bad and the Ugly," in \emph{Proc. IEEE Conf. on Comput. Vis. Pattern Recognit.},  Honolulu, Hawaii, July 2017, pp.~3077-3086.
	
	\bibitem{IEEEnips13:Frome}
	A.~Frome, G.~S.~Corrado, J.~Shlens, \emph{et al.}, ``DeViSE: A deep visual-semantic embedding model," Advances in \emph{Neural Inf. Process Syst.}, Nevada, US, Dec. 2013, pp.~2121-2129.
	
	\bibitem{akata2015label}
	Z.~Akata, F.~Perronnin, Z.~Harchaoui, \emph{et al.}, ``Label-embedding for image classification," \emph{IEEE Trans. Pattern Anal. Mach. Intell.},vol.~38, no.~7, pp.~1425-1438, Oct. 2015.
	
	\bibitem{IEEEnips15:Romera-Paredes}
	B.~Romera-Paredes and P.~H.~S~Torr, ``An embarrassingly simple approach to zero-shot learning," in \emph{Proc. Int. Conf. Mach. Learn.}, Lille, France, July 2015, pp.~2152-2161.
	
	\bibitem{IEEEcvpr16:Xian}
	Y.~Q.~Xian, Z.~Akata, G.~Sharma, \emph{et al.}, ``Latent embeddings for zero-shot classification," in \emph{Proc. IEEE Conf. on Comput. Vis. Pattern Recognit.}, Las Vegas, USA, June 2016, pp.~69-77.
	
	\bibitem{IEEEcvpr16:Changpinyo}
	S.~Changpinyo, W.~L.~Chao, B.~Gong, \emph{et al.}, ``Synthesized Classifiers for Zero-Shot Learning," in \emph{Proc. IEEE Conf. on Comput. Vis. Pattern Recognit.}, Las Vegas, USA, June 2016, pp.~5327-5336.
	
	\bibitem{IEEEnips13:Socher}
	R.~Socher, M.~Ganjoo, C.~D.~Manning, \emph{et al.}, ``Zero-shot learning through cross-modal transfer," Advances in \emph{Neural Inf. Process. Syst.}, Nevada, US, Dec. 2013, pp.~935-943.
	
	\bibitem{IEEEcvpr17:Kodirov}
	E.~Kodirov, T.~Xiang, and S.~Gong, ``Semantic Autoencoder for Zero-Shot Learning," in \emph{Proc. IEEE Conf. on Comput. Vis. Pattern Recognit.},  Honolulu, Hawaii, July 2017, pp.~4447-4456.
	
	\bibitem{iccv15:Zhang}
	Z.~Zhang, V.~Saligrama, ``Zero-shot learning via semantic similarity embedding," in \emph{Proc. IEEE Int. Conf. on Comput. Vis.}, Santiago, Chile, Dec. 2015, pp.~4166-4174.
	
	\bibitem{jiang2018learning}
	H.~Jiang, R.~Wang, S.~Shan, X.~Chen, ``Learning class prototypes via structure alignment for zero-shot recognition," in \emph{Proc. Eur. Conf. on Comput. Vis.}, Munich, Germany, Sept. 2018, pp.~118-134.
	
	\bibitem{IEEEiccv15:Kodirov}
	E.~Kodirov, T.~Xiang, Z.~Fu, \emph{et al.}, ``Unsupervised domain adaptation for zero-shot learning," in \emph{Proc. IEEE Conf. on Comput. Vis. Pattern Recognit.}, Santiago, Chile, Dec. 2015, pp.~2452-2460.
	
	\bibitem{ECML15:Shigeto}
	Y.~Shigeto, I.~Suzuki, K.~Hara, \emph{et al.}, ``Ridge Regression, Hubness, and Zero-Shot Learning," in \emph{Eur. Conf. Mach. Learn.}, Porto, Portugal, Sept. 2015, pp.135-151.
	
	\bibitem{IEEEcvpr17:Zhang}
	L.~Zhang, T.~Xiang, and S.~Gong, ``Learning a Deep Embedding Model for Zero-Shot Learning," in \emph{Proc. IEEE Conf. on Comput. Vis. Pattern Recognit.},  Honolulu, Hawaii, July 2017, pp.~3010-3019.
	
	\bibitem{zhao2017zero}
	B.~Zhao, B.~Wu, T.~Wu, \emph{et al.}, ``Zero-shot learning posed as a missing data problem," in \emph{Proc. IEEE Int. Conf. on Comput. Vis.}, Venice, Italy, Oct. 2017, pp.~2616-2622.
	
	\bibitem{wang2016relational}
	D.~Wang, Y.~Li, Y.~Lin, \emph{et al.}, ``Relational knowledge transfer for zero-shot learning," in Thirtieth \emph{AAAI Conf. Art. Intell.},  Phoenix, USA, Feb. 2016.
	
	\bibitem{IEEEcvpr18:Zhu}
	Y.~Zhu, M.~Elhoseiny, B.~Liu, X.~Peng, \emph{et al.}, ``A generative adversarial approach for zero-shot learning from noisy texts,"  in \emph{Proc. IEEE Conf. on Comput. Vis. Pattern Recognit.}, Salt Lake, USA, June 2018, pp.~899-907.
	
	\bibitem{felix2018multi}
	R.~Felix, V.~BG~Kumar, I.~Reid, \emph{et al.}, ``Multi-modal cycle-consistent generalized zero-shot learning," in \emph{Eur. Conf. Mach. Learn.}, Munich, Germany, Sept. 2018, pp.~21-37.
	
	\bibitem{chen2018zero}
	L.~Chen, H.~Zhang, J.~Xiao, \emph{et al.}, ``Zero-shot visual recognition using semantics-preserving adversarial embedding networks," in \emph{Proc. IEEE Conf. on Comput. Vis. Pattern Recognit.}, Salt Lake, USA, June 2018, pp.~1043-1052.
	
	\bibitem{chao2016empirical}
	W.~Chao, S.~Changpinyo, B.~Gong, \emph{et al.}, ``An empirical study and analysis of generalized zero-shot learning for object recognition in the wild," in \emph{Proc. Eur. Conf. on Comput. Vis.}, Amsterdam, The Netherlands, Oct. 2016, pp.~ 52-68.
	
	\bibitem{goodfellow2014generative}
	I.~Goodfellow, J.~Pouget-Abadie, M.~Mirza, \emph{et al.}, ``Generative adversarial nets," Advances in \emph{Neural Inf. Process. Syst.}, Montreal, Canada, Dec. 2014, pp.~2672-2680.
	
	\bibitem{IEEEcvpr18:Yang}
	F.~Yang, Y.~Sung, L.~Zhang, \emph{et al.}, ``Learning to Compare: Relation Network for Few-Shot Learning," in \emph{Proc. IEEE Conf. on Comput. Vis. Pattern Recognit.}, Salt Lake, USA, June 2018, pp.~1199-1208.
	
	\bibitem{zhang2019triple}
	H.~Zhang, Y.~Long, Y.~Guan, \emph{et al.}, ``Triple Verification Network for Generalized Zero-Shot Learning," \emph{IEEE Trans. Image. Process.}, vol.~28, no.~1, pp.~506-517, Jan. 2019.
	
	\bibitem{nair2010rectified}
	V.~Nair and G.~E.~Hinton, ``Rectified linear units improve restricted boltzmann machines," in \emph{Proc. Int. Conf. Mach. Learn.}, Haifa, Israel, June 2010, pp.~807-814.
	
	\bibitem{Yu2018zero}
	Y.~Yu, Z.~Ji, J.~Guo, \emph{et al.}, ``Zero-Shot learning via latent space encoding," \emph{IEEE Trans. Cybern.},  pp.~1-12, Jul. 2018.
	
	\bibitem{zhang2018sch}
	J.~Zhang, Y.~Peng and M.~Yuan, ``Sch-gan: Semi-supervised cross-modal hashing by generative adversarial network," \emph{IEEE Trans. Cybern.}, pp.~1-14, Sept. 2018.
	
	\bibitem{xian2018feature}
	Y.~Xian, T.~Lorenz, B.~Schiele, \emph{et al.}, ``Feature generating networks for zero-shot learning," in \emph{Proc. IEEE Conf. on Comput. Vis. Pattern Recognit.}, Salt Lake, USA, June 2018, pp.~5542-5551.
	
	\bibitem{atzmon2019adaptive}
	Y.~Atzmon and G.~Chechik, ``Adaptive Confidence Smoothing for Generalized Zero-Shot Learning," in \emph{Proc. IEEE Conf. on Comput. Vis. Pattern Recognit.}, Long Beach, USA, June 2019, pp.~11671-11680.
	
	\bibitem{bulent2019gradient}
	M.~Bulent Sariyildiz and R.~Gokberk Cinbis, ``Gradient Matching Generative Networks for Zero-Shot Learning," in \emph{Proc. IEEE Conf. on Comput. Vis. Pattern Recognit.}, Long Beach, USA, June 2019, pp.~2168-2178.
	
	\bibitem{shen2019scalable}
	F.~Shen, X.~Zhou, J.~Yu, \emph{et al.}, ``Scalable Zero-Shot Learning via Binary Visual-Semantic Embeddings," \emph{IEEE Trans. Image. Process.}, vol.~28, no.~7, pp.~3662 - 3674, Jul. 2019.
	
\end{thebibliography}
\end{document}